%% file: main.tex
\documentclass{article}
\usepackage{acra}
\usepackage[utf8]{inputenc}

\usepackage{graphicx}
\usepackage{subcaption}
\usepackage{enumitem}
\usepackage{booktabs}
\usepackage{gensymb}
\usepackage{tabularx}
\usepackage{amsmath}
\usepackage{comment}
\usepackage[capitalise]{cleveref}
\usepackage{multirow}
\usepackage{subcaption}
\usepackage{balance}

\usepackage[font=footnotesize]{caption}

\graphicspath{ {Images/} }

\usepackage{pgfplots}
\usetikzlibrary{external, positioning}
\usepackage{pgfplots}
\pgfplotsset{compat=1.9}
\usepgfplotslibrary{statistics}
\usetikzlibrary{shapes}
\usetikzlibrary{er}
\usepgfplotslibrary{colorbrewer}

\title{Comparing Subjective Perceptions of Robot-to-Human Handover Trajectories}

\author{Alexander Calvert \\ Monash University, Australia \\  acal0003@student.monash.edu
\And Wesley P. Chan \\ Monash University, Australia \\ wesley.chan@monash.edu 
\AND Tin Tran \\ Monash University, Australia \\ trung.tran@monash.edu
\And Sara Sheikholeslami \\ University of British Columbia, Canada \\ ssheikho@mail.ubc.ca
\AND Rhys Newbury \\ Monash University, Australia \\ Australian National University, Australia \\
rhys.newbury@monash.edu
\And Akansel Cosgun \\ Deakin University, Australia \\ akan.cosgun@deakin.edu.au
\AND Elizabeth Croft \\ University of Victoria, Canada \\ ecroft@uvic.ca
}

\begin{document}

\maketitle

\begin{abstract}
Robots must move legibly around people for safety reasons, especially for tasks where physical contact is possible. One such task is handovers, which requires implicit communication on where and when physical contact (object transfer) occurs. In this work, we study whether the trajectory model used by a robot during the reaching phase affects the subjective perceptions of receivers for robot-to-human handovers. We conducted a user study where 32 participants were handed over three objects with four trajectory models: three were versions of a minimum jerk trajectory, and one was an ellipse-fitting-based trajectory. The start position of the handover was fixed for all trajectories, and the end position was allowed to vary randomly around a fixed position by $\pm3$cm in all axis. The user study found no significant differences among the handover trajectories in survey questions relating to safety, predictability, naturalness, and other subjective metrics. While these results seemingly reject the hypothesis that the trajectory affects human perceptions of a handover, it prompts future research to investigate the effect of other variables, such as robot speed, object transfer position, object orientation at the transfer point, and explicit communication signals such as gaze and speech.
\end{abstract}

\section{Introduction}

Over the last few decades, robots have become more prevalent in various applications, including manufacturing~\cite{Tan2009}, healthcare~\cite{Kyrarini2021}, and home-care~\cite{Portugal2015}. Their use is projected to increase over the coming decade~\cite{roadmap2022}, requiring significant research and investment to ensure that robots can effectively integrate into human environments. One vital task that service robots need to perform frequently is object handovers with humans. An object handover is a collaborative task and involves an agent (the giver) handing an object over to another agent (the receiver). Since handovers are an essential part of our daily lives, robot-human handovers need to be a smooth, efficient, effective, and an enjoyable experience for humans.

In human-robot handovers, the human worker's selection of anticipatory actions has a temporal dependency on the robot's actions and is based on predictions of the future state of the robot's motion ~\cite{Dragan2013}. Such collaboration requires \textit{legible} (intent-expressive) coordination of the specific behavior of human-robot handovers. As the robot reaches out to handover an object, the collaborator should be able to tell early on and reach out to meet the robot's hand. 

Changing the style of the robot's motion to match the task it is performing also improves perception~\cite{Zhou2018}. As expectations strongly influence perception, robots need to perform handover motions to match the expectations and preferences of the human interacting with the robot~\cite{Bestick2018}. For example, a human would expect the robot to be confident while opening a door but cautious when handling a fragile object, resulting in two different motion styles.

Researchers investigating the reaching motion used by robots have found that human-like handover behavior is preferred~\cite{Shibata1997}, and can result in quicker and more efficient handovers~\cite{Huber2008}. Several human-like reaching motion models have been proposed and used in object handover research, including the minimum jerk model ~\cite{Hogan1982}, the decoupled minimum jerk model ~\cite{Huber2009} and more recently, the elliptical model~\cite{Sheikholeslami2018}. In our previous work, we compared these models to several datasets of human-giver reaching trajectories to investigate which model best fits the reaching motion of a human, finding that the elliptical model was more human-like compared to the minimum jerk models \cite{Sheikholeslami2018,Sheikholeslami2021}. However, these models are yet to be compared via a robot-human handover task. Hence, this paper aims to compare these models via a user study to establish which motion model is preferred by a human agent collaborating with a robotic system.

\begin{figure}[h!]
    \centering
    \includegraphics[width=0.93\linewidth]{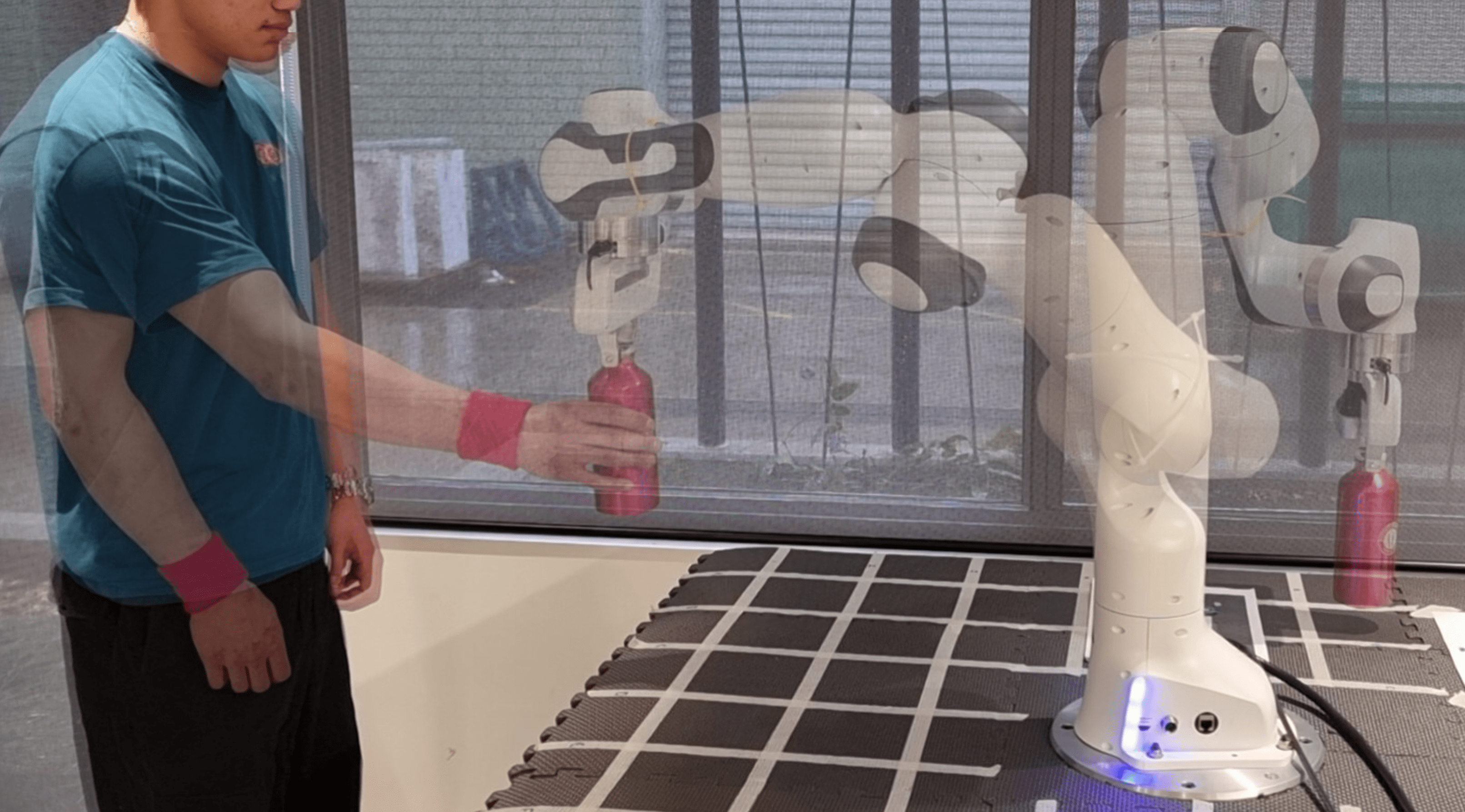}
    \caption{We compare the subjective experience of robot-to-human handovers for a range of motions.}
    \label{fig:intro_pic}
\end{figure}

\section{Related Works}

\subsection{Communication in Human-Robot Handovers}
A survey by ~\citeauthor{Ortenzi2020}~\shortcite{Ortenzi2020} found that only a minority of human-robot handover works consider communication cues. However, communication cues are critical in initiating and co-coordinating joint actions such as object handover ~\cite{Strabala2013}. These cues are often non-verbal and are modeled after human behavior. 

Several works have found that human-like gaze behaviors in robots can improve the timing of handovers~\cite{Moon2014,Zheng2015}. Gaze has also been shown to convey intent before a handover begins~\cite{Strabala2013}. ~\citeauthor{admoni2014deliberate}~\shortcite{admoni2014deliberate} decreased the speed of the handovers until the human gaze was drawn back to the robot, which increased the conscious perception of the communication cues. In ~\citeauthor{grigore2013joint}~\shortcite{grigore2013joint}, the authors integrated both head orientation and eye gaze into the robot's decision-making and showed that this significantly increases the success rate of robot-to-human handovers. The use of body gestures~\cite{Cakmak2011} and the initial pose of the robot~\cite{pan2018exploration} can improve the subjective experience of the handover.

\begin{figure*}[h!]
\begin{subfigure}{.24\textwidth}
  \centering
  \includegraphics[width=.8\linewidth]{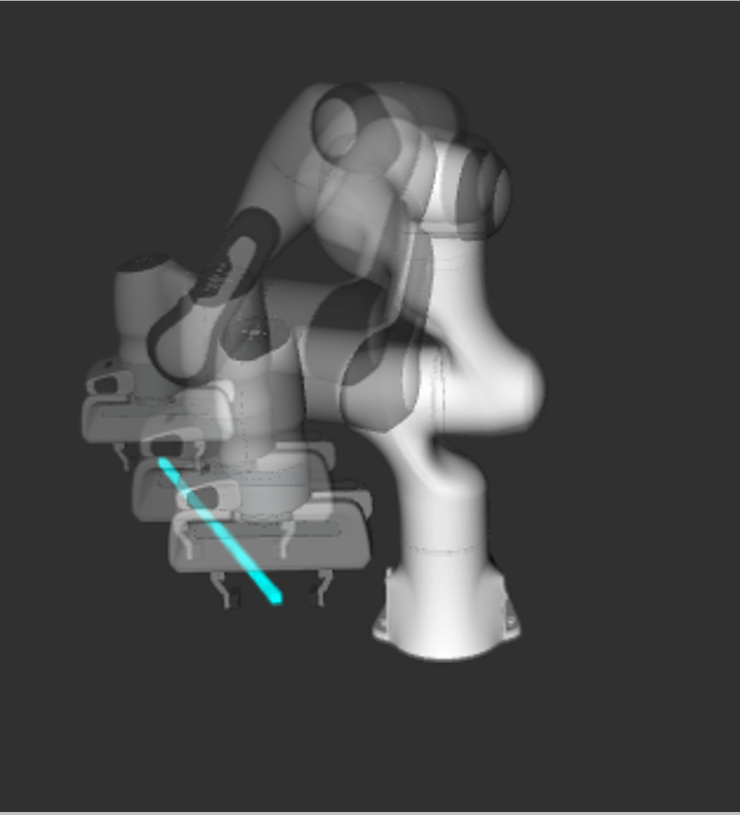}
  \caption{Minimum Jerk}
\end{subfigure}%
\begin{subfigure}{.24\textwidth}
  \centering
  \includegraphics[width=.8\linewidth]{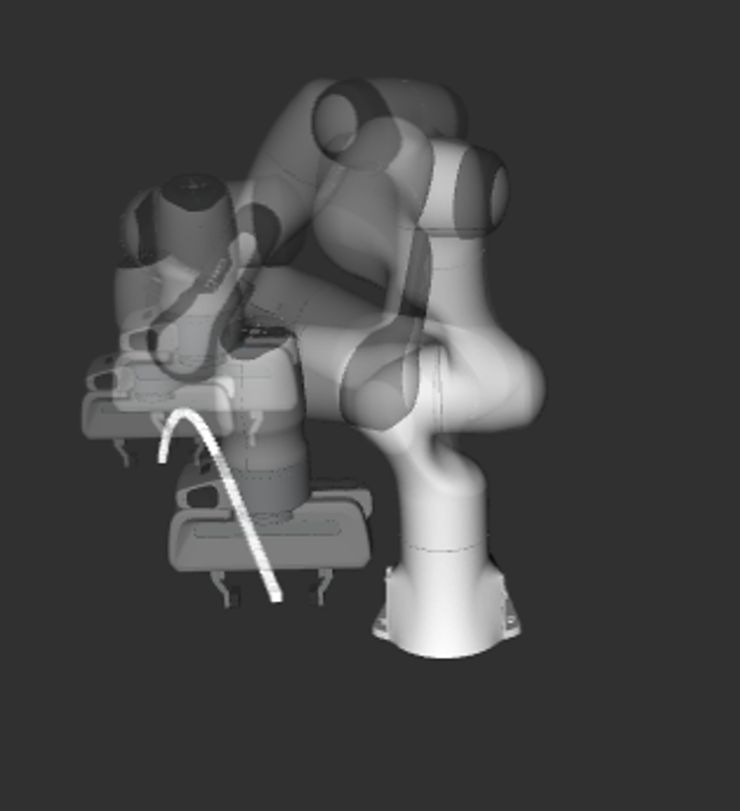}
  \caption{Decoupled Minimum Jerk}
\end{subfigure}%
\begin{subfigure}{.24\textwidth}
  \centering
  \includegraphics[width=.8\linewidth]{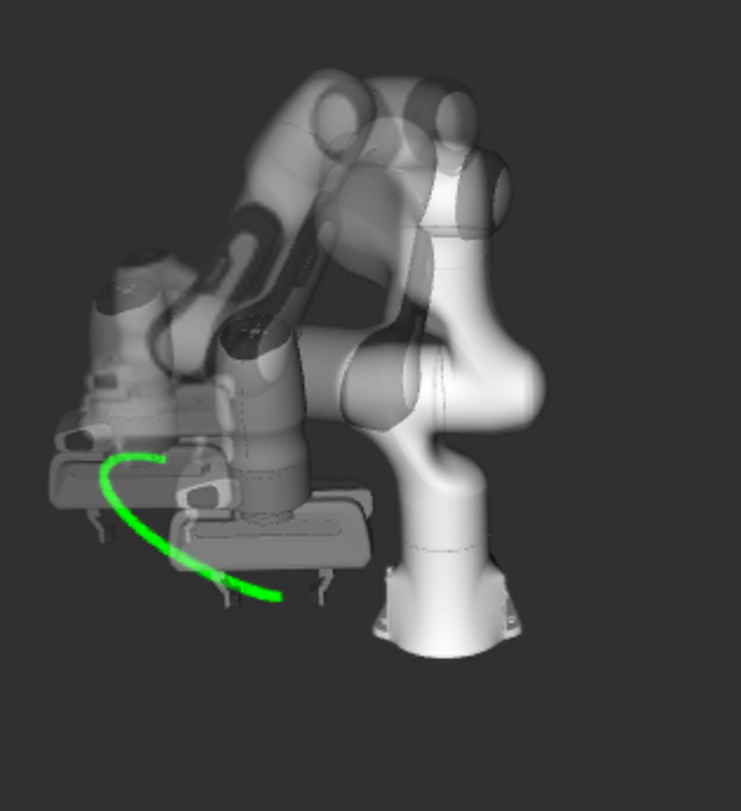}
  \caption{Elliptical Trajectory}
\end{subfigure}%
\begin{subfigure}{.24\textwidth}
  \centering
  \includegraphics[width=.8\linewidth]{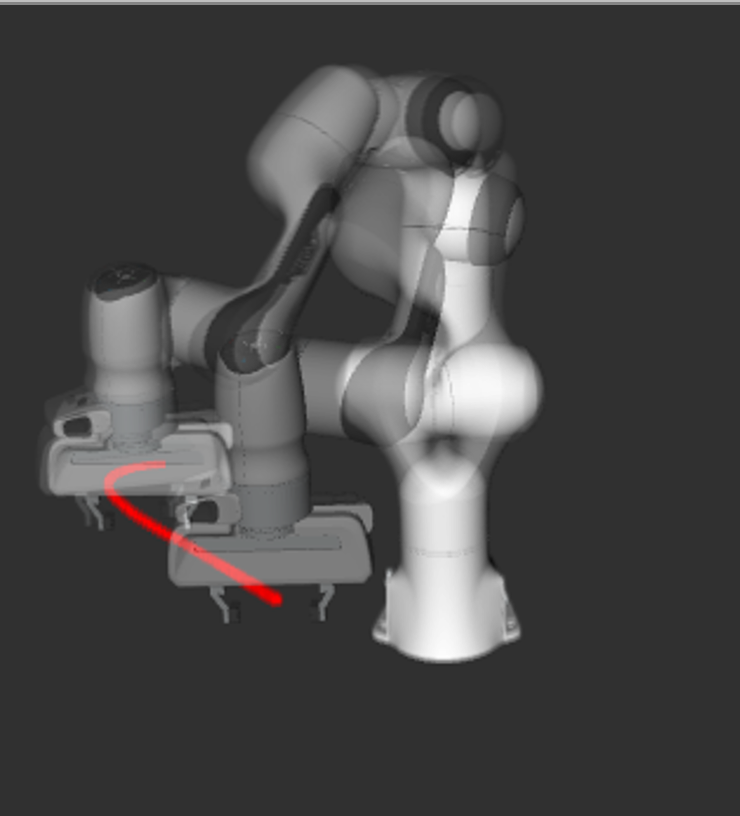}
  \caption{Slanted Decoupled Minimum Jerk}
\end{subfigure}
\caption{The four handover trajectories used in this study}
\label{fig:fig}
\end{figure*}

\subsection{Handover Motions}
The motion of the robot arm during the handover has also been shown to significantly affect subjective user experience during object handovers~\cite{Shibata1997}. Research has shown that humans can perceive the robot as safer and more pleasant when using human-like trajectories~\cite{Shibata1997,Huber2008}. In addition, human-like reaching motions can enable the prediction of the handover location, resulting in more timely and efficient handovers~\cite{Glasauer2010}. 

Given that object handovers can happen at any time, and with varying start and end positions, the reaching motion employed by a robot must be generated efficiently, requiring a model which can be computed in real-time. This requirement limits the number of models which are suitable for object handover. 

The Minimum Jerk trajectory model was first proposed by Hogan in the early 1980s~\cite{Hogan1982} and has been used extensively to model human handover motion, including in recent works such as~\citeauthor{Pan2019}~\shortcite{Pan2019}. The Minimum Jerk model assumes that the path taken during a handover is straight; however, it was later shown that human handover motions follow a curved path ~\cite{Huber2009}. To allow for a curved path,~\citeauthor{Huber2009}~\shortcite{Huber2009} proposed the Decoupled Minimum Jerk model. This model creates a curved path by decoupling the reaching motion into two separate Minimum Jerk trajectories. More recently, ~\citeauthor{Sheikholeslami2018}~\shortcite{Sheikholeslami2018} proposed an elliptical motion model that is shown to yield a good fit to human motion in seated single-person pick-and-place tasks. Research has also investigated whether this model fits the human motion in unconstrained object handover, showing that the elliptical model fits the motion accurately~\cite{Sheikholeslami2021}.

\section{Trajectory Generation Models}

The scope of this work is limited to the trajectory that the robot arm takes between the point the object is first grasped from the table and the object transfer point, which is fixed in all experiments. We compare four trajectory models, each described in this section.

\subsection{Minimum Jerk}

The Minimum Jerk model hypothesizes that human arm motion between two points in space minimizes the jerk over the entire path, which implies a straight line path between the start and the endpoint in euclidean space~\cite{Hogan1982}. Given position as a function of time $r(t)$, jerk is the third derivative of position, $\dddot{r}(t)$. The smoothness of the path, $S$, can be measured by

\begin{equation}
S = \int_{0}^{t_f} \dddot{r}(t)^2dt
\end{equation}
where $t_{f}$ is the duration of the motion. The Minimum Jerk trajectory is the trajectory r(t) that minimizes the smoothness S. The solution takes on the form of a fifth degree polynomial~\cite{Hogan1982}:
\begin{equation}
r(t) = a_0 + a_1t + a_2t^2 + a_3t^3 + a_4t^4 + a_5t^5
\end{equation}
where the coefficients $a_i$ are determined by the boundary conditions, i.e., the position, velocity, and acceleration at the start and end points of the trajectory. 

\subsection{Decoupled Minimum Jerk}

The Decoupled Minimum Jerk model can be specified by two Minimum Jerk trajectories~\cite{Huber2009}, one for the $XY$-plane motion and the other for the $Z$ axis motion, each with a different duration where the $Z$ axis is along the local vertical direction. The Decoupled Minimum Jerk trajectories are described by:
\begin{equation}
r_z(t) = a_{0z} + a_{1z}t + a_{2z}t^2 + a_{3z}t^3 + a_{4z}t^4 + a_{5z}t^5,
\end{equation}
\begin{equation}
\begin{aligned}
r_{xy}(t) = a_{0xy} + a_{1xy}t + a_{2xy}t^2 + \\ a_{3xy}t^3 + a_{4xy}t^4 + a_{5xy}t^5,
\end{aligned}
\end{equation}
where $r_z(t)$ is the trajectory in the $Z$ direction, with duration $t_z$, and $r_{xy}(t)$ is the trajectory in the $XY$-plane, with duration $t_{xy}$. Coefficients $a_{iz}$ and $a_{ixy}$ are determined by boundary conditions and the desired duration of each component of the motion. The Decoupled Minimum Jerk trajectory is on a plane orthogonal to the $XY$-plane, and has a curved path in 3D space. 

\subsection{Elliptical Model}
It has recently been empirically shown that an elliptical curve achieves a better fit to human reaching motions for both unconstrained~\cite{Sheikholeslami2021} and constrained~\cite{Sheikholeslami2018} handovers. 

The equation for an ellipse can be expressed in parametric form, parameterized with $\theta$, as:

\begin{equation}
\label{ellipticalparametric}
\begin{bmatrix} x(\theta)\\y(\theta)\end{bmatrix} = \begin{bmatrix} x_c\\y_c\end{bmatrix} + \begin{bmatrix} cos(\tau) & -sin(\tau)\\sin(\tau) & cos(\tau)\end{bmatrix} \begin{bmatrix} asec(\theta)\\btan(\theta)\end{bmatrix}
\end{equation}

where $x_c$, $y_c$ is the centre of the ellipse, $\tau$ is the roll angle and $a,b$ are the semi-major and semi-minor axes respectively. 


The dataset by~\citeauthor{Chan2020handover}~\shortcite{Chan2020handover} is utilized to fit the ellipse models. A subset of trajectories (76 out of 1195) was used, selecting only right-handed handovers and trajectories that fit the elliptical model well. Parameters in \cref{ellipticalparametric} were computed from the mean values of the trajectories in the dataset, along with the starting and ending points of the handover. The starting point is placed above the table top, to the right of the robot arm, while the ending point is set at the mean end position of the subset of trajectories.

Ellipses arise as second-degree curves generated by the intersection of a plane and a cone. Therefore, to justify an elliptical fit to the reach data, the reach trajectory must be planar~\cite{Sheikholeslami2021}. The same subset of data was analyzed to find the normal vector for each trajectory to determine the best fit plane. We then used the mean normal vector and the known starting position to define the plane on which the robot's motion would be generated. 


An elliptical trajectory is then generated by specifying the dependence of $\theta$ on time. As the dataset provided by ~\citeauthor{Chan2020handover}~\shortcite{Chan2020handover} includes unconstrained handovers with non-zero initial velocity and acceleration, fitting a model to this dataset would result in the robot jerking rapidly at the start of the reaching motion. To prevent this, we utilized an additional dataset~\cite{Sheikholeslami2018} of handover reaching motions with zero initial velocity and acceleration. We normalized time and $\theta$ to be in the range [0,1] and fit a sigmoid function to the mean $\theta$ from this dataset to define an elliptical trajectory.

\subsection{Slanted Decoupled Minimum Jerk}

We included a new different type of trajectory, generated by specifying a Decoupled Minimum Jerk trajectory and rotating it to be on the same plane as the Elliptical Motion Model used. Adding this additional trajectory allows a more direct comparison of the Decoupled Minimum Jerk trajectory to the Elliptical Motion Model.

\section{User Study Design}
\label{sec:user_study}

\subsection{Hardware}

We use a 7-DoF Franka Emika Panda robotic arm controlled via Robot Operating System (ROS). An OAK-D camera operating at 60Hz was utilized to record the participant whilst they completed each handover, with the data recorded used for post-analysis to determine the location of the user's hand during the handover. Participants wore a red band on their right wrist, allowing easy hand tracking.

\subsection{Task Description}

To hand over the objects to the receiver, the robot first moves from a home position to a fixed position above the objects. It then picks up an object and moves to a fixed starting position. The robot then utilizes a handover motion model to hand the object over to the receiver. Once the robot reaches the object transfer point, it waits for the user to take the object, pausing for 1 second before moving back to the home position.

Research has shown that a handover typically occurs in approximately 1.2 seconds~\cite{Chan2020handover}. Due to constraints on the joint velocity of the robot, the fastest time the robot could move from the starting position to the handover location was four seconds. This was the duration used for all motion models.

Another important aspect of the robot's motion is the handover location, the position at which the robot finishes its reaching motion. Some researchers have used adaptive controllers which adjust for the position of the receiver's hand, while others use a fixed position~\cite{Moon2014}. We chose to utilize a fixed position; however, the handover location was allowed to vary randomly around this fixed position by up to $\pm3$ centimeters in all three axes. This slight variation in the end position was added to minimize the chances that the receiver would memorize and anticipate the handover location after they became familiar with the system.

\begin{figure}[t]
    \centering
    \includegraphics[width=0.6\linewidth]{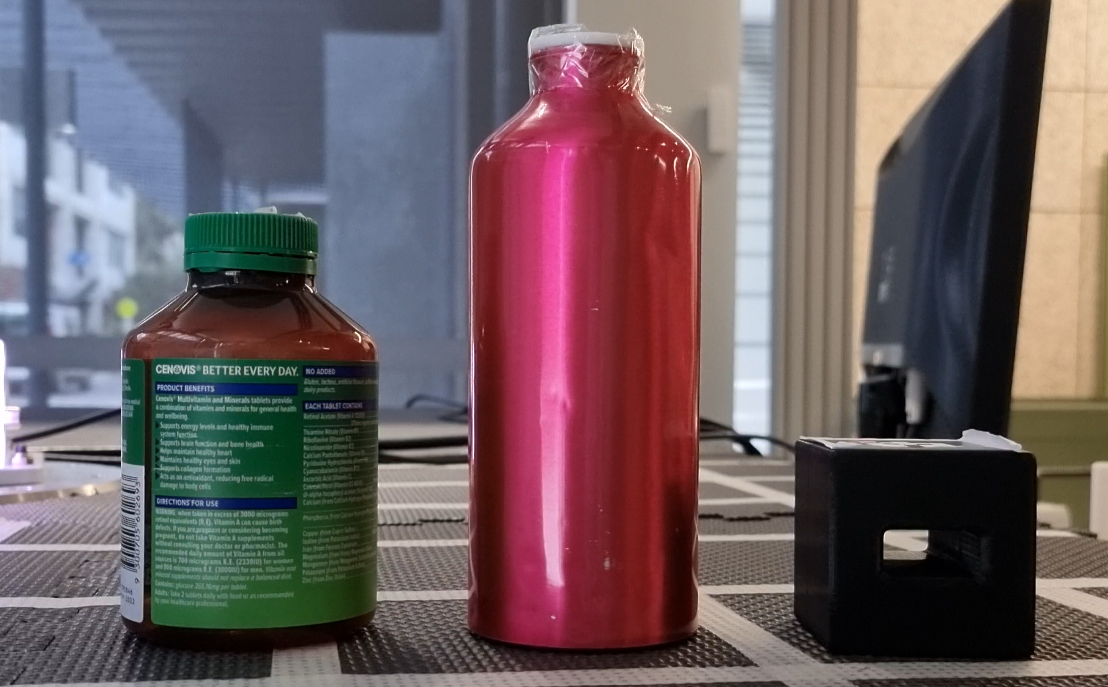}
    \caption{Objects used in the user study. Left: vitamin bottle, Middle: water bottle, Right: cube}
    \label{objects}
\end{figure}

\subsection{Independent Variable}
We utilized a single independent variable, which was the reaching motion used by the robot.

Four different motions were used:
\begin{itemize}
    \item Minimum Jerk (MJ)
    \item Decoupled Minimum Jerk (DMJ)
    \item Slanted Decoupled Minimum Jerk (SDMJ)
    \item Elliptical (E)
\end{itemize}

The order that each participant experienced the conditions was counterbalanced to alleviate possible ordering effects.

\subsection{Objects}

For each condition, the participants received three objects (water bottle, vitamin bottle, and cube) from the robot, as shown in \cref{objects}. Given that the experiment focuses purely on the motion of the handover trajectory, the starting position of each object was fixed relative to the robot base. The external forces and torques on the end effector were utilized to detect when the receiver had grasped the object.

\subsection{Participant Allocation}

We recruited 32 participants ($20$ male and $12$ female) aged from $19-28$ (M=22.7, SD=2.09)\footnote{This study has been approved by the Monash University Human Research Ethics Committee (Application ID: 27499)}. The participants were not compensated for their time. Seven of the participants had previously worked with robotics, two of the participants had completed similar studies, while the rest of the participants reported either not seeing a robot or only seeing commercially available robots.

\subsection{Procedure}
The study was conducted in a university laboratory, with an experimenter supervising the experiment. Upon arrival, participants were provided an explanatory statement to read and a consent form to sign. They then completed a demographic survey. Next, the experiment was outlined to the participant, and they then completed three practice handovers with the robot. During practice, the handover motion used by the robot for each handover was randomly selected, with no motion being used for more than one practice handover.

Participants were instructed to start each handover with their reaching hand by their side. The participant was then required to complete 12 successful handovers with the robot. After each condition, the participant completed a survey regarding their experience with that condition before moving to the next. At the end of all of the conditions, the user completed an additional post-experiment survey where they compared each of the conditions and had the opportunity to provide additional comments.

\subsection{Objective Metrics}

Four objective measures were recorded and analyzed during this study: (1) the percentage of successful handovers, (2) the start time of the human reach (Reach Time), (3) the middle time of reach  (Middle Time), and (4) the time that the robot's gripper begins to release the object (Release Time). Each objective metric is defined more precisely below, and all time-based metrics are measured from when the robot starts the reaching motion. The Reach and Middle times were calculated from the recorded OAK-D videos of each handover. We used color segmentation techniques to determine the location of the participant's hand at each step.

\begin{enumerate}
    \item Success: The percentage of handovers which were successful.
    \item Reach Time: The time at which the receiver begins to move their hand towards the object to grab it.
    \item Middle Time: The time at which the receivers hand crosses a fixed pixel location as they move towards the object. This position was fixed for all participants, and represented the half-way position in pixel space, between the object transfer point and where the human was standing. 
    \item Release Time: The time at which the robot begins to release the object from the gripper.
\end{enumerate}

All time metrics were measured in seconds. We included the Reach Time and Middle Time metrics due to the discovery made by~\citeauthor{Moon2014}~\shortcite{Moon2014} that there was no difference in Release Time for different gaze behavior in object handovers, but there was a significant difference in Reach Time. Therefore, we have included an additional metric, Middle Time, to investigate whether there are significant differences in the reaction times across the entire motion of the receiver during the handover.

\subsection{Subjective Metrics}
After each condition, participants were asked a series of questions to establish which handover motion they subjectively preferred. Additionally, according to the recommendations made by~\citeauthor{Ortenzi2020}~\shortcite{Ortenzi2020}, we included several questions to evaluate the fluency of the interaction, trust, and working alliance of the human-robot interaction. All questions were asked on a 7-point Likert scale. The participants' questions are listed in \cref{tab:questions}.

\noindent\begin{table}[h!]
    \centering  
    \caption{User Study Survey Questions.}
    \begin{tabularx}{\linewidth}{|X|}
        \toprule
        \textbf{Human-Robot Fluency} \\
        Q1: The human-robot team worked fluently together \\
        Q2: The robot contributed to the fluency of the interaction\\
         \textbf{Trust in Robot}\\
        Q3: I trusted the robot to do the right thing at the right time \\
        Q4: The robot was trustworthy \\
        \textbf{Safety} \\
        Q5: I felt safe during the handover \\
        \textbf{Predictability}\\
        Q6: I understand what the robot’s goals are\\
        Q7: The robot and I are working towards mutually agreed upon goals \\
        \textbf{Natural Motion}\\
        Q8: The motion was natural \\
        Q9: The motion was human-like \\
        Q10: I was comfortable with the handover motion \\
        \bottomrule
    \end{tabularx}
    \label{tab:questions}
    \vspace{-0.3cm}
\end{table}

After participants completed all four conditions, they were asked which condition they preferred and whether they perceived any differences between the conditions, with the option to provide written feedback.

\begin{figure*}[h!]
    \centering
    \input{boxplot}
    \caption{Subjective Participant Responses. E, DMJ, MJ, SDMJ represent Elliptical, Decoupled Minimum Jerk, Minimum Jerk, and Slanted Decoupled Minimum Jerk, respectively.}
    \label{fig:boxplot}
\end{figure*}
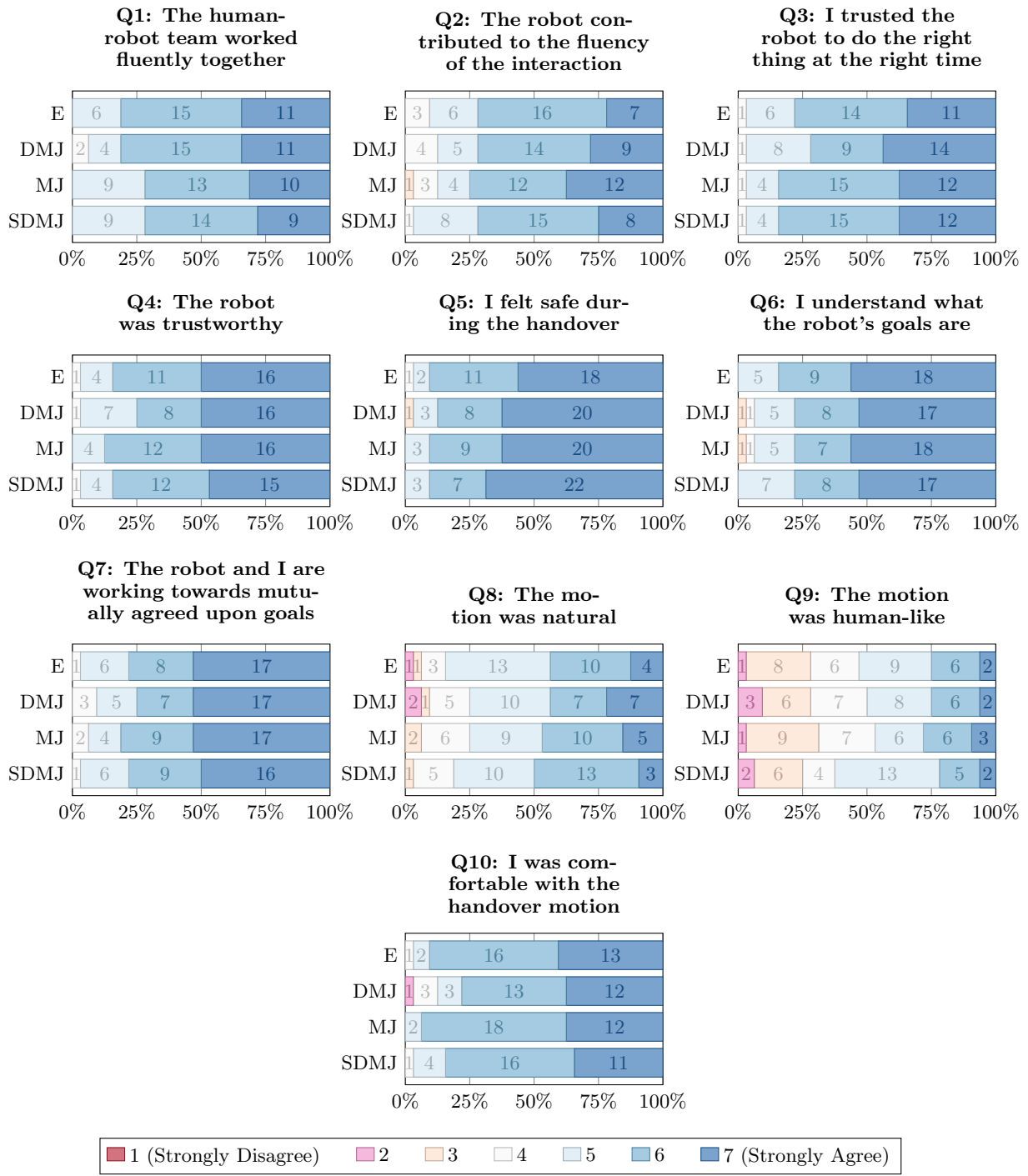

\section{Results}
\label{sec:results}

\begin{table*}[t!]
\centering
\caption{Objective metric results. We observe no significant difference between any of the trajectories for the objective metrics. E, DMJ, MJ, SDMJ represent Elliptical, Decoupled Minimum Jerk, Minimum Jerk, and Slanted Decoupled Minimum Jerk, respectively. Bold numbers indicate the motion/s with the lowest mean time value (in seconds) for each objective metric}
\label{tab:objresults}
\begin{tabular}{|p{2.5cm}|c|c|c|c|c|c|c|c|c|c|}
\hline
 &  \multicolumn{2}{c|}{E} & \multicolumn{2}{c|}{DMJ} & \multicolumn{2}{c|}{MJ} & \multicolumn{2}{c|}{SDMJ} & \multirow{2}{*}{F(1,63)} & \multirow{2}{*}{p} \\ \cline{1-9} 
 & $\mu$ & $\sigma$ & $\mu$ & $\sigma$ & $\mu$ & $\sigma$ & $\mu$ & $\sigma$ & & \\ \hline
Reach Time & 4.28 & 0.59 & \textbf{4.15} & 0.49 & \textbf{4.15} & 0.56 & 4.27 & 0.57 & 0.53 & 0.66 \\ \hline
Middle Time & 5.02 & 0.55 & 4.96 & 0.48 & \textbf{4.95} & 0.50 & 5.03 & 0.48 & 0.20 & 0.90 \\ \hline
Release Time & 5.95 & 0.55 & 5.87 & 0.73 & 5.89 & 0.72 & \textbf{5.84} & 0.53 & 0.15 & 0.93 \\ \hline
\end{tabular}
\end{table*}

\begin{table*}[h!]
\centering
\caption{User Responses to the Survey Questions shown in Table~\ref{tab:questions}. E, DMJ, MJ, SDMJ represent Elliptical, Decoupled Minimum Jerk, Minimum Jerk, and Slanted Decoupled Minimum Jerk, respectively. Bold numbers indicate the motion/s with the highest mean for each question}
\label{tab:subresults}
\begin{tabular}{|l|c|c|c|c|c|c|c|c|c|c|}
\hline
 &  \multicolumn{2}{c|}{E} & \multicolumn{2}{c|}{DMJ} & \multicolumn{2}{c|}{MJ} & \multicolumn{2}{c|}{SDMJ} & \multirow{2}{*}{F(1,63)} & \multirow{2}{*}{p} \\ \cline{1-9}
 & $\mu$ & $\sigma$ & $\mu$ & $\sigma$ & $\mu$ & $\sigma$ & $\mu$ & $\sigma$ & & \\ \hline

Q1 & \textbf{6.16} & 0.72 & 6.09 & 0.86 & 6.03 & 0.78 & 6.00 & 0.76 & 0.25 & 0.86 \\ \hline

Q2 & 5.84 & 0.88 & 5.88 & 0.98 & \textbf{5.97} & 1.09 & 5.94 & 0.80 & 0.12 & 0.95 \\ \hline

Q3 & 6.09 & 0.82 & 6.13 & 0.91 & \textbf{6.19} & 0.78 & \textbf{6.19} & 0.78 & 0.10 & 0.96 \\ \hline

Q4 & 6.31 & 0.82 & 6.22 & 0.91 & \textbf{6.38} & 0.71 & 6.28 & 0.81 & 0.20 & 0.89 \\ \hline

Q5 & 6.44 & 0.76 & 6.44 & 0.91 & 6.53 & 0.67 & \textbf{6.59} & 0.67 & 0.33 & 0.81 \\ \hline

Q6 & \textbf{6.41} & 0.76 & 6.22 & 1.04 & 6.25 & 1.05 & 6.31 & 0.82 & 0.26 & 0.86 \\ \hline

Q7 & \textbf{6.28} & 0.89 & 6.19 & 1.03 & \textbf{6.28} & 0.92 & 6.25 & 0.88 & 0.07 & 0.97 \\ \hline

Q8 & 5.31 & 1.12 & 5.25 & 1.39 & 5.31 & 1.15 & \textbf{5.38} & 0.98 & 0.06 & 0.98 \\ \hline

Q9 & 4.53 & 1.32 & 4.44 & 1.41 & 4.50 & 1.41 & \textbf{4.59} & 1.32 & 0.07 & 0.97 \\ \hline

Q10 & 6.28 & 0.73 & 5.97 & 1.18 & \textbf{6.31} & 0.59 & 6.16 & 0.77 & 1.09 & 0.36 \\ \hline
\end{tabular}
\end{table*}

\subsection{Objective Metrics}
Out of the 385 handovers completed, only one handover was classified as failing. The participant took the object with their left hand rather than their right hand, re-attempting this handover immediately afterward. This produced a success rate of 99.7\%.

The results for the remaining objective metrics are shown in \cref{tab:objresults}. We conducted a one-way ANOVA analysis with a significance level of $\alpha = 0.05$, which revealed no significant differences in the three time-based metrics.

\subsection{Subjective Metrics}

The survey results are shown in \cref{fig:boxplot}. A one-way ANOVA was conducted on the Likert scale subjective questions, returning no significant difference between the conditions for any of the ten questions. The statistics for these questions are summarised in \cref{tab:subresults}. Participants commonly mentioned that they could not perceive any difference between the conditions.

\section{Discussion}

Our results show no significant difference between any of the motion models in our task for any subjective or objective metrics measured, indicating that participants could not perceive significant differences. Many participants also indicated in their verbal responses that they could not identify changes in the robot motion between conditions. A couple of participants could determine that there were two planes on which the end-effector moved (one for DMJ and MJ and another for E and SDMJ), but they could not tell the difference between the two motions that traveled on each plane.

An interesting observation was the substantially lower average scores (across all motions) for two of the subjective Likert questions compared to the rest. The lower-scored questions were ``The motion was human-like" (Q9) and ``The motion was natural" (Q8), which both aimed to determine which motion the participants perceived as more human-like. The lower average scores for these two questions compared to the rest may indicate that participants did not consider any of the motions very human-like. 

The highest scoring question (Q5) was in relation to safety, which suggests that users felt comfortable with the robotic platform and the robot's motions. This is likely due to the speed of the robot and the reliability of our system to perform handovers. ``The robot was trustworthy" (Q4) and ``I understand what the robot’s goals are" (Q6) were also rated highly, suggesting that the users were trusting of the robot and understood the robot's goals. 

A factor that may explain our results is the configuration of our robotic platform, which can be seen in \cref{fig:intro_pic}. Common robotic platforms, including the Franka Emika Panda robotic arm used in our experiments, tend not to be very human-like when mounted in a tabletop configuration. Using a more human-like platform, such as a humanoid robot or a horizontally-mounted robotic configuration~\cite{disney} could help humans perceive the motions as more natural or human-like. Furthermore, we only consider the trajectory of the end-effector.

However, efficiently configuring the robot arm coordinated joint trajectories to closely match human whole arm configurations while respecting the kinematics of the manipulator could help users perceive the motion in a more human-like manner. Additionally, the robot joints move quite rapidly near the start of the handover motion to ensure the end-effector follows the required path, which may distract the participant from observing the robot's exact handover motion more closely.

These results also indicate that the handover motion utilized on non-humanoid robotic systems may not be consequential to successfully perform object handovers. Participants subjectively rated all four conditions highly for fluency, comfort, trustworthiness, working alliance and safety, indicating that whilst the handover motions were not that well perceived as human-like or natural, they were satisfied with their interactions with the robot. We think it would be valuable to perform the same study again, however, varying the handover scenarios by changing the object transfer point, human position, and arm speed. This could help expose the differences between the trajectories to the users and expose a stronger need for legible motions. 


\section{Conclusion and Future Work}
\label{sec:conclusion}


We compared the subjective perception of different trajectories on robot-to-human handovers. For the trajectories tested, all of which were selected for their human-inspired nature, our results did not find significant differences in successful robot-to-human handover interactions. This finding is in contrast to an earlier hypothesis on human-to-human object handovers~\cite{Sheikholeslami2018,Sheikholeslami2021} which posited that the elliptical model would enable more fluent and efficient object handovers as compared to the other human-like motion models.

An obvious extension of this work is to investigate the effect of the other factors on the subjective perception of humans, such as the robot speed and a varying object transfer point of a handover. Future research can also examine the effect of combining human-like trajectories with additional explicit communication methods, such as gaze~\cite{Moon2014,Zheng2015}, speech~\cite{chao2010turn} gestures~\cite{kwan2020gesture} and augmented reality markers~\cite{newbury2021}. The motion of the robot base, in addition to the robotic arm, can also be considered~\cite{he2022go}. Furthermore, we believe looking at the effect of the subjective perception of human-like trajectories on human-to-robot handovers~\cite{Rosenberge2020,yang2021reactive} could help provide exciting insights. Another interesting research avenue is investigating whether the conclusions reached for robot-to-human handovers are also valid for human-to-robot handovers.

\balance
\bibliographystyle{named.bst}
\bibliography{refs.bib}
\end{document}

%% file: boxplot.tex
\begin{tikzpicture}
    \definecolor{one}{RGB}{178,24,43}
    \definecolor{two}{RGB}{239,138,198}
    \definecolor{three}{RGB}{253,219,199}
    \definecolor{four}{RGB}{247, 247, 247}
    \definecolor{five}{RGB}{209,229,240}
    \definecolor{six}{RGB}{103, 169, 207}
    \definecolor{seven}{RGB}{33,102,172}

    \pgfplotsset{
      /pgfplots/bar  cycle  list/.style={/pgfplots/cycle  list={%
            {one!75!black,fill=one!60!white,mark=none},%
            {two!75!black,fill=two!60!white,mark=none},%
            {three!75!black,fill=three!60!white,mark=none},%
            {four!75!black,fill=four!60!white,mark=none},%
            {five!75!black,fill=five!60!white,mark=none},%
            {six!75!black,fill=six!60!white,mark=none},%
            {seven!75!black,fill=seven!60!white,mark=none}%
         }
      },
    }
    
    \tikzstyle{every node}=[font=\small]

    
        
        
        
        
        
    

    \begin{axis}[
        name=mainplot,
        xbar stacked,
        title style = {text width=4cm, align = center},
        title = \textbf{Q1: The human-robot team worked fluently together},
        nodes near coords,
        bar width=0.8,
        width = 0.32\textwidth,
        height = 0.225\textwidth,
        xmin = 0, xmax = 32,
        enlarge y limits={abs=10pt},
        ytick={0,1,2,3},
        yticklabels={SDMJ, MJ, DMJ, E},  
        xtick={0,8,16,24,32}, 
        xticklabels={0\%,25\%,50\%,75\%,100\%},         
        legend style={at={(-10,-0.20)}, anchor=north, legend columns=-1, /tikz/every even column/.append style={column sep=0.5cm}},
    ]
    
        \addplot coordinates
        {(0,3) (0,2) (0,1) (0,0)};
        
        \addplot coordinates
        {(0,3) (0,2) (0,1) (0,0)};
                
        \addplot coordinates
        {(0,3) (0,2) (0,1) (0,0)};
                
        \addplot coordinates
        {(0,3) (2,2) (0,1) (0,0)};

        \addplot coordinates
        {(6,3) (4,2) (9,1) (9,0)};

        \addplot coordinates
        {(15,3) (15,2) (13,1) (14,0)};

        \addplot coordinates
        {(11,3) (11,2) (10,1) (9,0)};

    \end{axis}
    \begin{axis}[
        name=secondplot,
        title style = {text width=4cm, align = center},
        title=\textbf{Q2: The robot contributed to the fluency of the interaction},
        at={(mainplot.north east)},
        xshift=1.2cm,
        anchor=north west,    
        xbar stacked,
        nodes near coords,
        bar width=0.8,
        width = 0.32\textwidth,
        height = 0.225\textwidth,
        xmin = 0, xmax = 32,
        enlarge y limits={abs=10pt},
        ytick={0,1,2,3},
        yticklabels={SDMJ, MJ, DMJ, E},  
        xtick={0,8,16,24,32}, 
        xticklabels={0\%,25\%,50\%,75\%,100\%},       
        legend style={at={(0.5,-0.20)}, anchor=north, legend columns=-1, /tikz/every even column/.append style={column sep=0.5cm}},
    ]
    
        \addplot coordinates
        {(0,3) (0,2) (0,1) (0,0)};
        
        \addplot coordinates
        {(0,3) (0,2) (0,1) (0,0)};
                
        \addplot coordinates
        {(0,3) (0,2) (1,1) (0,0)};
                
        \addplot coordinates
        {(3,3) (4,2) (3,1) (1,0)};

        \addplot coordinates
        {(6,3) (5,2) (4,1) (8,0)};

        \addplot coordinates
        {(16,3) (14,2) (12,1) (15,0)};

        \addplot coordinates
        {(7,3) (9,2) (12,1) (8,0)};
        
    \end{axis}
    \begin{axis}[
        name=thirdplot,
        title style = {text width=4cm, align = center},
        title=\textbf{Q3: I trusted the robot to do the right thing at the right time},
        at={(secondplot.north east)},
        xshift=1.2cm,
        anchor=north west,      
        xbar stacked,
        nodes near coords,
        bar width=0.8,
        width = 0.32\textwidth,
        height = 0.225\textwidth,
        xmin = 0, xmax = 32,
        enlarge y limits={abs=10pt},
        ytick={0,1,2,3},
        yticklabels={SDMJ, MJ, DMJ, E},  
        xtick={0,8,16,24,32}, 
        xticklabels={0\%,25\%,50\%,75\%,100\%},      
        legend style={at={(0.5,-0.20)}, anchor=north, legend columns=-1, /tikz/every even column/.append style={column sep=0.5cm}},
    ]
    
        \addplot coordinates
        {(0,3) (0,2) (0,1) (0,0)};
        
        \addplot coordinates
        {(0,3) (0,2) (0,1) (0,0)};
                
        \addplot coordinates
        {(0,3) (0,2) (0,1) (0,0)};
                
        \addplot coordinates
        {(1,3) (1,2) (1,1) (1,0)};

        \addplot coordinates
        {(6,3) (8,2) (4,1) (4,0)};

        \addplot coordinates
        {(14,3) (9,2) (15,1) (15,0)};

        \addplot coordinates
        {(11,3) (14,2) (12,1) (12,0)};
        
    \end{axis}
    \begin{axis}[
        name=fourplot,
        at={(mainplot.below south west)},
        title style = {text width=4cm, align = center},
        title=\textbf{Q4: The robot was trustworthy},
        yshift=-1.3cm,
        anchor=north west,
        xbar stacked,
        nodes near coords,
        bar width=0.8,
        width = 0.32\textwidth,
        height = 0.225\textwidth,
       xmin = 0, xmax = 32,
        enlarge y limits={abs=10pt},
        ytick={0,1,2,3},
        yticklabels={SDMJ, MJ, DMJ, E},  
        xtick={0,8,16,24,32}, 
        xticklabels={0\%,25\%,50\%,75\%,100\%},      
        legend style={at={(0.5,-0.20)}, anchor=north, legend columns=-1, /tikz/every even column/.append style={column sep=0.5cm}},
    ]
        \addplot coordinates
        {(0,3) (0,2) (0,1) (0,0)};
        
        \addplot coordinates
        {(0,3) (0,2) (0,1) (0,0)};
                
        \addplot coordinates
        {(0,3) (0,2) (0,1) (0,0)};
                
        \addplot coordinates
        {(1,3) (1,2) (0,1) (1,0)};

        \addplot coordinates
        {(4,3) (7,2) (4,1) (4,0)};

        \addplot coordinates
        {(11,3) (8,2) (12,1) (12,0)};

        \addplot coordinates
        {(16,3) (16,2) (16,1) (15,0)};

    \end{axis}
    \begin{axis}[
        name=fiveplot,
        title style = {text width=4cm, align = center},
        title=\textbf{Q5: I felt safe during the handover},
        at={(secondplot.below south west)},
        yshift=-1.3cm,
        anchor=north west,    
        xbar stacked,
        nodes near coords,
        bar width=0.8,
        width = 0.32\textwidth,
        height = 0.225\textwidth,
        xmin = 0, xmax = 32,
        enlarge y limits={abs=10pt},
        ytick={0,1,2,3},
        yticklabels={SDMJ, MJ, DMJ, E},  
        xtick={0,8,16,24,32}, 
        xticklabels={0\%,25\%,50\%,75\%,100\%},     
        legend style={at={(0.5,-0.35)}, xshift=-3cm, anchor=north, legend columns=-1, /tikz/every even column/.append style={column sep=0.5cm}},
    ]
    
        \addplot coordinates
        {(0,3) (0,2) (0,1) (0,0)};
        
        \addplot coordinates
        {(0,3) (0,2) (0,1) (0,0)};
                
        \addplot coordinates
        {(0,3) (1,2) (0,1) (0,0)};
                
        \addplot coordinates
        {(1,3) (0,2) (0,1) (0,0)};

        \addplot coordinates
        {(2,3) (3,2) (3,1) (3,0)};

        \addplot coordinates
        {(11,3) (8,2) (9,1) (7,0)};

        \addplot coordinates
        {(18,3) (20,2) (20,1) (22,0)};

        
    \end{axis}
    
        \begin{axis}[
        name=sixplot,
        title style = {text width=4cm, align = center},
        title=\textbf{Q6: I understand what the robot's goals are},
        at={(thirdplot.below south west)},
        yshift=-1.3cm,
        anchor=north west,    
        xbar stacked,
        nodes near coords,
        bar width=0.8,
        width = 0.32\textwidth,
        height = 0.225\textwidth,
        xmin = 0, xmax = 32,
        enlarge y limits={abs=10pt},
        ytick={0,1,2,3},
        yticklabels={SDMJ, MJ, DMJ, E},  
        xtick={0,8,16,24,32}, 
        xticklabels={0\%,25\%,50\%,75\%,100\%},      
        legend style={at={(0.5,-0.35)}, xshift=-3cm, anchor=north, legend columns=-1, /tikz/every even column/.append style={column sep=0.5cm}},
    ]
    
        \addplot coordinates
        {(0,3) (0,2) (0,1) (0,0)};
        
        \addplot coordinates
        {(0,3) (0,2) (0,1) (0,0)};
                
        \addplot coordinates
        {(0,3) (1,2) (1,1) (0,0)};
                
        \addplot coordinates
        {(0,3) (1,2) (1,1) (0,0)};

        \addplot coordinates
        {(5,3) (5,2) (5,1) (7,0)};

        \addplot coordinates
        {(9,3) (8,2) (7,1) (8,0)};

        \addplot coordinates
        {(18,3) (17,2) (18,1) (17,0)};

        
    \end{axis}

        \begin{axis}[
        at={(fourplot.below south west)},
        title style = {text width=4cm, align = center},
        title=\textbf{Q7: The robot and I are working towards mutually agreed upon goals},
        yshift=-1.7cm,
        anchor=north west,
        xbar stacked,
        nodes near coords,
        bar width=0.8,
        width = 0.32\textwidth,
        height = 0.225\textwidth,
        xmin = 0, xmax = 32,
        enlarge y limits={abs=10pt},
        ytick={0,1,2,3},
        yticklabels={SDMJ, MJ, DMJ, E},  
        xtick={0,8,16,24,32}, 
        xticklabels={0\%,25\%,50\%,75\%,100\%},      
        legend style={at={(0.5,-0.20)}, anchor=north, legend columns=-1, /tikz/every even column/.append style={column sep=0.5cm}},
    ]
        \addplot coordinates
        {(0,3) (0,2) (0,1) (0,0)};
        
        \addplot coordinates
        {(0,3) (0,2) (0,1) (0,0)};
                
        \addplot coordinates
        {(0,3) (0,2) (0,1) (0,0)};
                
        \addplot coordinates
        {(1,3) (3,2) (2,1) (1,0)};

        \addplot coordinates
        {(6,3) (5,2) (4,1) (6,0)};

        \addplot coordinates
        {(8,3) (7,2) (9,1) (9,0)};

        \addplot coordinates
        {(17,3) (17,2) (17,1) (16,0)};

    \end{axis}
    \begin{axis}[
        name=bottomplot,
        title style = {text width=4cm, align = center},
        title=\textbf{Q8: The motion was natural},
        at={(fiveplot.below south west)},
        yshift=-1.7cm,
        anchor=north west,    
        xbar stacked,
        nodes near coords,
        bar width=0.8,
        width = 0.32\textwidth,
        height = 0.225\textwidth,
        xmin = 0, xmax = 32,
        enlarge y limits={abs=10pt},
        ytick={0,1,2,3},
        yticklabels={SDMJ, MJ, DMJ, E},  
        xtick={0,8,16,24,32}, 
        xticklabels={0\%,25\%,50\%,75\%,100\%},       
        legend style={at={(0.5,-0.35)}, xshift=-3cm, anchor=north, legend columns=-1, /tikz/every even column/.append style={column sep=0.5cm}},
    ]
    
        \addplot coordinates
        {(0,3) (0,2) (0,1) (0,0)};
        
        \addplot coordinates
        {(1,3) (2,2) (0,1) (0,0)};
                
        \addplot coordinates
        {(1,3) (1,2) (2,1) (1,0)};
                
        \addplot coordinates
        {(3,3) (5,2) (6,1) (5,0)};

        \addplot coordinates
        {(13,3) (10,2) (9,1) (10,0)};

        \addplot coordinates
        {(10,3) (7,2) (10,1) (13,0)};

        \addplot coordinates
        {(4,3) (7,2) (5,1) (3,0)};

        
    \end{axis}
    
        \begin{axis}[
        title style = {text width=4cm, align = center},
        title=\textbf{Q9: The motion was human-like},
        at={(sixplot.below south west)},
        yshift=-1.7cm,
        anchor=north west,    
        xbar stacked,
        nodes near coords,
        bar width=0.8,
        width = 0.32\textwidth,
        height = 0.225\textwidth,
        xmin = 0, xmax = 32,
        enlarge y limits={abs=10pt},
        ytick={0,1,2,3},
        yticklabels={SDMJ, MJ, DMJ, E},  
        xtick={0,8,16,24,32}, 
        xticklabels={0\%,25\%,50\%,75\%,100\%},    
        legend style={at={(0.5,-0.35)}, xshift=-3cm, anchor=north, legend columns=-1, /tikz/every even column/.append style={column sep=0.5cm}},
    ]
    
       \addplot coordinates
        {(0,3) (0,2) (0,1) (0,0)};
        
        \addplot coordinates
        {(1,3) (3,2) (1,1) (2,0)};
                
        \addplot coordinates
        {(8,3) (6,2) (9,1) (6,0)};
                
        \addplot coordinates
        {(6,3) (7,2) (7,1) (4,0)};

        \addplot coordinates
        {(9,3) (8,2) (6,1) (13,0)};

        \addplot coordinates
        {(6,3) (6,2) (6,1) (5,0)};

        \addplot coordinates
        {(2,3) (2,2) (3,1) (2,0)};

        
    \end{axis}
        \begin{axis}[
        title style = {text width=4cm, align = center},
        title=\textbf{Q10: I was comfortable with the handover motion},
        at={(bottomplot.below south west)},
        yshift=-1.7cm,
        anchor=north west,    
        xbar stacked,
        nodes near coords,
        bar width=0.8,
        width = 0.32\textwidth,
        height = 0.225\textwidth,
        xmin = 0, xmax = 32,
        enlarge y limits={abs=10pt},
        ytick={0,1,2,3},
        yticklabels={SDMJ, MJ, DMJ, E},  
        xtick={0,8,16,24,32}, 
        xticklabels={0\%,25\%,50\%,75\%,100\%},       
        legend style={at={(1.1,-0.35)}, xshift=-3cm, anchor=north, legend columns=-1, /tikz/every even column/.append style={column sep=0.5cm}},
    ]
    
        \addplot coordinates
        {(0,3) (0,2) (0,1) (0,0)};
        
        \addplot coordinates
        {(0,3) (1,2) (0,1) (0,0)};
                
        \addplot coordinates
        {(0,3) (0,2) (0,1) (0,0)};
                
        \addplot coordinates
        {(1,3) (3,2) (0,1) (1,0)};

        \addplot coordinates
        {(2,3) (3,2) (2,1) (4,0)};

        \addplot coordinates
        {(16,3) (13,2) (18,1) (16,0)};

        \addplot coordinates
        {(13,3) (12,2) (12,1) (11,0)};

        \legend{1 (Strongly Disagree), 2, 3, 4, 5, 6, 7 (Strongly Agree)}
        
    \end{axis}

    
        
        
        
        
        

\end{tikzpicture}